\documentclass{article}

\PassOptionsToPackage{numbers}{natbib}

\usepackage{graphicx}
\usepackage{wrapfig}
\usepackage[preprint]{neurips_2026}
\usepackage{amsmath}
\usepackage{booktabs}
\usepackage{multirow}
\usepackage[table]{xcolor}
\usepackage{algorithm}
\usepackage{algpseudocode}
\usepackage{amsmath}
\usepackage{amssymb}

\usepackage[utf8]{inputenc} 
\usepackage[T1]{fontenc}    
\usepackage{hyperref}       
\usepackage{url}            
\usepackage{booktabs}       
\usepackage{amsfonts}       
\usepackage{nicefrac}       
\usepackage{microtype}      
\usepackage{xcolor}         

\newtheorem{proposition}{Proposition}

\title{ConQuR: Corner Aligned Activation Quantization via Optimized Rotations for LLMs}

\author{%
  Chayne Thrash \\
  Department of Computer Science\\
  Vanderbilt University\\
  \texttt{chayne.thrash@vanderbilt.edu} \\
  \And
  Ali Abbasi \\
  Department of Computer Science\\
  Vanderbilt University\\
  \texttt{ali.abbasi@vanderbilt.edu} \\
  \And
  Soheil Kolouri \\
  Department of Computer Science\\
  Vanderbilt University\\
  \texttt{soheil.kolouri@vanderbilt.edu} \\
}

\begin{document}

\maketitle

\begin{abstract}

Large language models (LLMs) are costly to deploy due to their large memory footprint and high inference cost. Weight-activation quantization can reduce these costs, but low-bit activation quantization remains difficult because activation outliers induce large quantization error. Recent rotation-based methods address this by applying orthogonal transformations that redistribute activation magnitude across dimensions, but existing approaches either require expensive end-to-end rotation training or rely on stored activation corpora, introducing significant compute or storage overhead. We propose a lightweight post-training rotation calibration method for LLM activation quantization. Our method learns orthogonal rotations that align normalized activations with the corners of an inscribed hypercube, encouraging activation energy to be distributed more evenly across dimensions. This objective admits an efficient closed-form update via the orthogonal Procrustes problem, avoiding gradient-based optimization over the orthogonal group. We further introduce an online calibration procedure that updates rotations as calibration samples are processed, eliminating the need to store activations on disk and allowing rotations to adapt to quantized activation distributions during calibration. Experiments on Llama-2 and Llama-3 models from 3B to 70B parameters show that our method achieves competitive or improved performance across perplexity benchmarks and common sense reasoning tasks while avoiding both costly end-to-end training and large offline activation storage.
  
\end{abstract}

\section{Introduction}

Large language models (LLMs)~\cite{llama_2,llama_3,gpt} have achieved remarkable performance across a wide range of domains, including natural language understanding and generation, multi-step reasoning, code synthesis, and mathematical problem solving~\cite{gpt,cot,palm,codex}. However, these capabilities are enabled by models with billions of parameters, making inference expensive in terms of memory footprint, latency, and energy consumption. These costs limit the deployment of LLMs in practical settings, particularly on resource-constrained hardware or in applications requiring high-throughput inference.

A variety of compression techniques have been developed to reduce the cost of deploying large models. These include pruning methods, which remove unimportant weights or structures from the model~\cite{sparsegpt,wanda,llm_pruner}; singular value decomposition (SVD) and low-rank approximation methods, which replace dense weight matrices with compact low-rank factors~\cite{svd_llm,asvd}; and knowledge distillation, which transfers the behavior of a large teacher model to a smaller student model~\cite{minillm}. Among these approaches, quantization has emerged as one of the most effective and widely used strategies for improving inference efficiency.

Quantization reduces the memory and compute requirements of LLMs by representing weights, activations, or KV-cache tensors using lower-precision numerical formats. Weight-only quantization methods have shown that LLM weights can often be compressed to very low bit-widths while preserving model quality~\cite{optq,awq}. However, quantizing activations remains substantially more challenging. A primary obstacle is the presence of large activation outliers, which expand the dynamic range of the quantizer and reduce the effective resolution assigned to the majority of activation values~\cite{smoothquant,llm_int8}. As a result, low-bit weight-activation quantization typically requires additional mechanisms to control or suppress activation outliers.

Several approaches have been proposed to address this challenge. Mixed-precision methods allocate higher precision to sensitive layers, channels, or activation groups~\cite{zeroquant,omniquant}. Smoothing-based methods use diagonal rescaling transformations to migrate activation outliers into the weights, thereby reducing the activation quantization range~\cite{smoothquant,outlier_suppression_plus}. More recently, rotation-based methods have emerged as a promising alternative for mitigating activation outliers. These methods apply orthogonal transformations to hidden states, spreading outlier dimensions across the feature space while preserving the model function by merging inverse rotations into adjacent linear layers~\cite{quarot,spinquant,dartquant, dfrot}. Unlike diagonal smoothing, rotations can redistribute activation energy across all dimensions, making them particularly attractive for low-bit activation quantization.

Existing rotation-based quantization methods typically follow one of two strategies: they either optimize rotations end-to-end using a task loss, which can require substantial compute, or they calibrate rotations using objectives that explicitly reshape activation distributions. While the latter avoids the cost of end-to-end fine-tuning, it often requires storing large collections of intermediate activations, leading to substantial storage overhead. Moreover, because these activations are collected before optimization, the calibration procedure is decoupled from the quantized forward pass that will be used at inference time.

In this work, we address these limitations with a rotation-based method for
improving LLM activation quantization by aligning normalized activations with
vertices of an inscribed hypercube. This objective encourages each activation
vector to distribute its magnitude more evenly across feature dimensions,
reducing the dominance of outlier coordinates that otherwise determine the
quantization scale. Building on the observation that rotation calibration can be
cast as an orthogonal Procrustes problem, our method optimizes this geometric
objective with lightweight closed-form updates. The resulting online mini-batch
procedure calibrates rotations from a small number of full sequences at a time,
without storing a large activation corpus, and can update rotations using the
quantized activations produced by the current model. This matches calibration
more closely to inference-time conditions while retaining the efficiency of
post-training rotation calibration.

\section{Related work}

\paragraph{Post-training quantization and activation outliers.}
Post-training quantization (PTQ) reduces the memory and inference cost of LLMs without expensive retraining. Weight-only methods such as GPTQ/OPTQ~\cite{optq}, AWQ~\cite{awq}, and SqueezeLLM~\cite{squeeze_llm} achieve strong compression by minimizing layer-wise reconstruction error or protecting activation-sensitive weights, while systems such as ZeroQuant combine weight and activation quantization in a full PTQ pipeline~\cite{zeroquant}. However, weight-only quantization leaves activations in floating point and therefore does not fully realize the efficiency benefits of low-precision inference. Low-bit weight-activation quantization is more challenging because LLM activations contain large, structured outliers that expand the quantization range and reduce effective resolution. Prior work addresses this with equivalent transformations: SmoothQuant migrates quantization difficulty from activations into weights using per-channel scaling~\cite{smoothquant}, Outlier Suppression+ combines shifting and scaling to reduce asymmetric activation outliers~\cite{outlier_suppression_plus}, and OmniQuant learns quantization parameters and equivalent transformations in a block-wise PTQ framework~\cite{omniquant}. 

\paragraph{Rotation-based quantization and incoherence processing.}
A closely related family of methods uses orthogonal transformations to distribute outliers across dimensions while preserving model outputs in full precision. QuIP introduced incoherence processing for LLM quantization, using randomized orthogonal transformations to make weight matrices more amenable to low-bit quantization and providing theoretical guarantees for 2-bit PTQ~\cite{quip}. QuIP\# and QTIP then extended this idea to low-bit vector quantization~\cite{quip_sharp,qtip}. Although these methods primarily target weight-only quantization, they established the utility of orthogonal transformations for reshaping tensor distributions before quantization.

Recent work has adapted this idea to activation quantization. QuaRot applies fixed Hadamard rotations throughout the Transformer to remove activation outliers, enabling 4-bit quantization of weights, activations, and KV caches with efficient inference kernels~\cite{quarot}. DuQuant combines rotations and permutations to distribute both normal and massive activation outliers across channels and blocks~\cite{duquant}. SpinQuant shows that the choice of rotation can significantly affect quantized accuracy and learns rotation matrices directly to improve downstream quantized performance~\cite{spinquant}. DartQuant further proposes distribution-aware rotational calibration, reducing the computational and memory burden of learning rotations while improving scalability to larger models~\cite{dartquant}. Most similar to our work, DFRot avoids the complexity of gradient descent on the orthogonal group by instead directly optimizing quantization error via closed form Procrustes updates~\cite{dfrot}. 

Our work builds on this line of rotation-based quantization methods. DartQuant~\cite{dartquant} calibrates
rotations using stored activation distributions, while DFRot~\cite{dfrot} uses
Procrustes-based optimization to directly reduce activation quantization error.
Rather than optimizing quantization error on a small fixed activation subset, we
optimize a hypercube-corner alignment objective and apply the resulting
Procrustes updates online over successive mini-batches of full sequences. This
avoids large activation storage and allows rotations to adapt under active
activation quantization to the distributions encountered during quantized
inference.

\section{Method}
\label{sec:method}

\begin{figure*}[!t]
    \centering
    \includegraphics[width=1.0\linewidth]{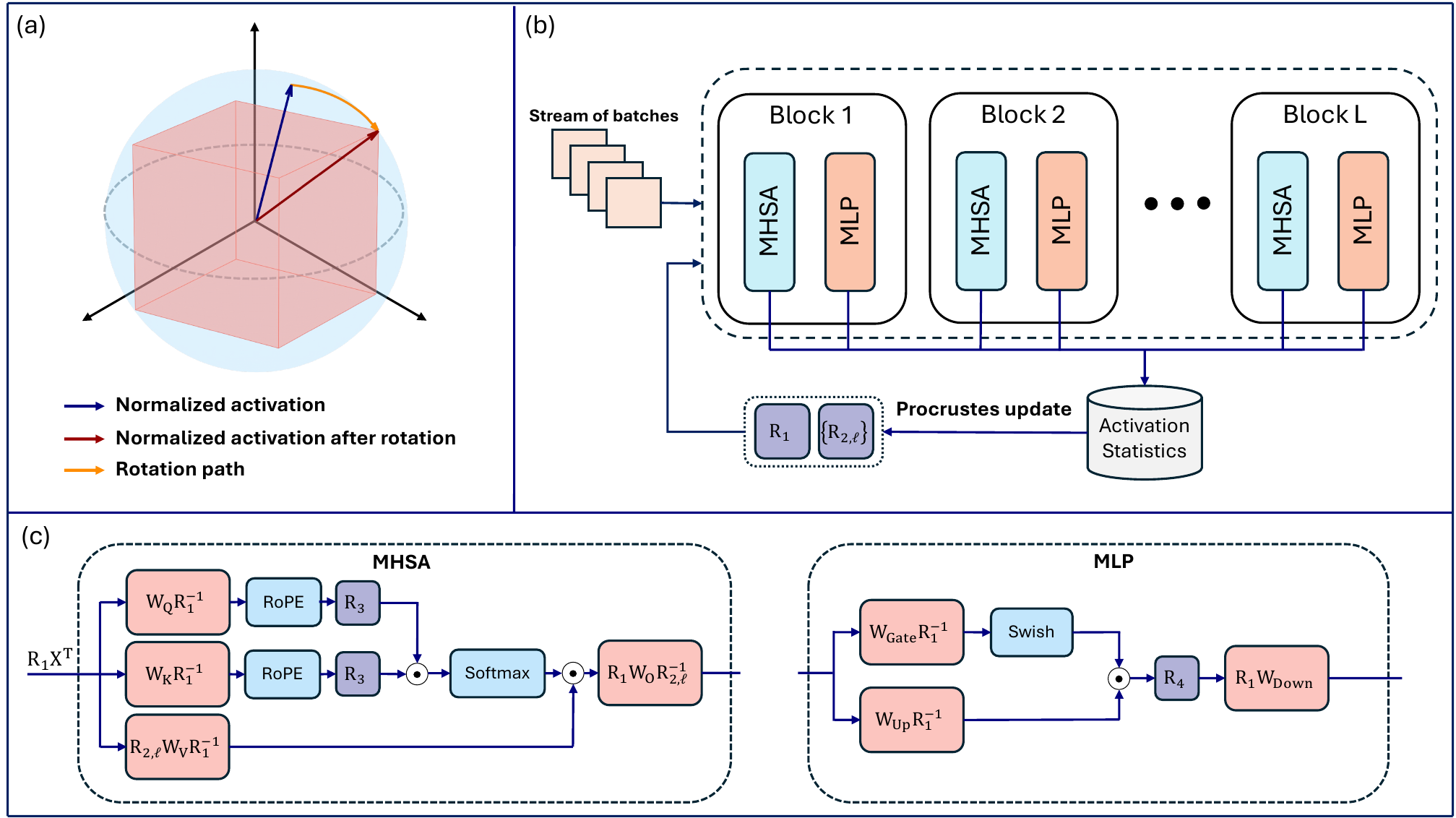}
    \caption{
Overview of the proposed rotation-based calibration method.
(a) Our method learns an orthogonal rotation that aligns normalized activation vectors with vertices of an inscribed hypercube, encouraging activation magnitude to be distributed more evenly.
(b) During calibration, activations are processed online in mini-batches with closed-form orthogonal Procrustes updates.
(c) At inference, learned rotations \(R_{1}, \{R_{2,\ell}\}_{l=1}^{L}\)are folded into linear layer weights and Hadamard rotations, \(R_3\) and \( R_4\), are applied online.
}
    \label{fig:main}
\end{figure*}

In this section, we present a rotation-based method for improving activation quantization by aligning activations with the corners of a hypercube. Figure~\ref{fig:main} provides an overview of the approach. An overview of how rotations are applied within the LLM can be seen in panel (c) and a full description is provided in Appendix \ref{rotation}. We begin by introducing our optimization objective and relate it directly to quantization error. Then, we show that this objective admits an orthogonal Procrustes solution. Finally, we describe how we apply this optimization within an online calibration algorithm. 


\subsection{Aligning activations with corners of a hyper-cube}
A central challenge in activation quantization for large language models is that the hidden state often uses coordinates unevenly: a small subset of dimensions may dominate the magnitude of the activation, while many others remain comparatively small. Such imbalance leads to inefficient use of the quantization grid, since a few large coordinates determine the scale while the remaining coordinates are quantized with unnecessarily coarse resolution. We address this issue by learning an orthogonal rotation that makes normalized activations more balanced across dimensions.

Let \(x^{i} \in \mathbb{R}^{d}\) denote an activation vector collected from the model, where \(i\) indexes activations across a calibration set. We optimize an orthogonal matrix \(R \in O(d)\) by solving

\begin{equation}\label{eq:objective}
    \min_{R \in O(d)} \sum_{i} \left \| R \frac{x^{i}}{\left \| x^{i}\right \|} - z^{i} \right \|_{2}^{2}
\end{equation}

where the target \(z^{i} \in \mathbb{R}^d\) is defined coordinatewise as

\begin{equation}
z^{i}_{j} = \frac{1}{\sqrt{d}} \text{sign}((R x^{i})_{j}).    
\end{equation}

The target \(z^{i}\) is a vertex of the hypercube inscribed in the unit hypersphere. Indeed, each coordinate of \(z^{i}\) has magnitude \(1 / \sqrt{d}\), so \(\left \|z^{i}\right\|_{2}=1\). Geometrically, the objective encourages the rotated activations to move towards a direction in which all coordinates have equal absolute magnitude. As no single dimension is allowed to dominate the representation, this optimization encourages an activation's features to be distributed more evenly across its dimensions.


\subsection{From quantization error to the corner objective}
\label{sec:theory}
For a uniform activation quantizer whose scale is determined by the
per-token dynamic range, the worst-case squared error is controlled by the
largest coordinate magnitude of the activation. For a normalized activation
$\tilde{x}$ and an orthogonal rotation $R$, the $\ell_2$ norm is fixed, so
rotation can only improve this bound by reducing
$\|R\tilde{x}\|_\infty$. The smallest possible value of
$\|y\|_\infty$ on the unit sphere is $1/\sqrt{d}$, achieved exactly when
all coordinates have equal magnitude, i.e., at the vertices of the hypercube
inscribed in the unit sphere.

Directly minimizing $\|R\tilde{x}\|_\infty$ is non-smooth and depends only
on the largest coordinate. We instead use a dense dual surrogate. Since
$\ell_1$ and $\ell_\infty$ are dual norms under Hölder's inequality, and
since $\|y\|_2=1$ is fixed, maximizing $\|y\|_1$ encourages the opposite
geometry of large $\|y\|_\infty$: it rewards spreading mass evenly across
coordinates. In particular, Cauchy--Schwarz gives
\[
\|y\|_1 \leq \sqrt{d}\|y\|_2=\sqrt{d},
\]
with equality if and only if
$|y_1|=\cdots=|y_d|=1/\sqrt{d}$. Thus, on the unit sphere, maximizing
$\|R\tilde{x}\|_1$ and minimizing $\|R\tilde{x}\|_\infty$ have the same
per-sample optima: the hypercube corners.

Our corner-distance objective is exactly this $\ell_1$ surrogate in
squared-distance form. For
$z^i=d^{-1/2}\operatorname{sign}(R\tilde{x}^i)$,
\[
\sum_i \|R\tilde{x}^i-z^i\|_2^2
=
2n-\frac{2}{\sqrt{d}}\sum_i \|R\tilde{x}^i\|_1 
\]
where \(n\) is the number of activations. Therefore, minimizing distance to the nearest hypercube corner is equivalent
to maximizing the average $\ell_1$ norm of the rotated normalized
activations. This gives a dense objective with the same ideal geometry as
the $\ell_\infty$ dynamic-range objective, while leading to the closed-form
Procrustes update described below. Appendix~\ref{app:quant_error} provides
the full derivation from symmetric quantization error to the
$\ell_\infty$ dynamic-range ratio, as well as the extension to asymmetric
zeropoint quantizers.

\subsection{Alternating orthogonal Procrustes updates}
\label{sec:procrustes}

The prior objective \ref{eq:objective} is non-smooth because the target \(z^{i}\) depends on the sign pattern of the rotated activation. Thus, we use an alternating procedure that updates the targets for a fixed rotation and then updates the rotation for the fixed targets. This yields a simple iterative algorithm in which each rotation update reduces to an orthogonal Procrustes problem.

Let $\tilde{x}^{i}=\frac{x^{i}}{\left\|x^{i}\right\|_{2}}$ denote the normalized activation vector. After fixing the targets, we can rewrite equation \ref{eq:objective} by stacking normalized activations and targets into matrices \(\tilde{X}, Z \in \mathbb{R}^{n \times d}\). The subproblem can then be written as

\[
\min_{R \in O(d)} \left\|\tilde{X} R^{\top} - Z \right\|_{F}^{2}
\]

This problem is exactly the orthogonal Procrustes problem and has a known closed form solution. Let

\[
C = Z^{\top} \tilde{X}
\]

with singular value decomposition

\[
C = U \Sigma V^{\top}.
\]

The optimal rotation is then given by

\[
R \leftarrow U V^{\top}.
\]

We summarize this entire procedures with the operation 
\[
R = \mathrm{OPU}(R,X),
\]
where OPU stands for orthogonal Procrustes update.

A useful feature of this formulation is that it avoids gradient-based optimization over the orthogonal group. Instead of parameterizing \(R\) through a differentiable surrogate and performing gradient-based optimization, each update is solved exactly for the current targets.

\begin{algorithm}[t]
\caption{Online calibration of shared and layerwise rotations}
\label{alg:online_calibration}
\begin{algorithmic}[1]
\Require calibration sequences $\mathcal{D} = \{s_i\}_{i=1}^N$, mini-batch size $B$, number of transformer blocks $L$, initial shared rotation $R_1 \in \mathbb{R}^{d \times d}$, initial block-diagonal rotations $\{R_{2,\ell}\}_{\ell=1}^L$ with $R_{2,\ell} \in \mathbb{R}^{n_h d_h \times n_h d_h}$
\For{mini-batch $\mathcal{B} \subset \mathcal{D}$ of size $B$}
    \State Initialize empty batch activations $\mathcal{X}_1$ associated with $R_1$
    \State Run the model on $\mathcal{B}$ layer by layer
    \For{$\ell = 1, \dots, L$}
        \State Collect self-attention and MLP input activations $X_{\mathrm{attn}}^{(\ell)}$, $X_{\mathrm{mlp}}^{(\ell)}$
        \State Accumulate $X_{\mathrm{attn}}^{(\ell)}$ and $X_{\mathrm{mlp}}^{(\ell)}$ into $\mathcal{X}_1$
        \State Collect $o\_proj$ input activations $X_o^{(\ell)}$
        \State $R_{2,\ell} \gets \mathrm{OPU}(R_{2,\ell}, X_o^{(\ell)})$
    \EndFor
    \State $R_1 \gets \mathrm{OPU}(R_1, \mathcal{X}_1)$
\EndFor
\State \Return $R_1$, $\{R_{2,\ell}\}_{\ell=1}^L$
\end{algorithmic}
\end{algorithm}

\subsection{Online mini-batch calibration}

We combine our Procrustes formulation with a simple online calibration algorithm. Instead of solving for the rotation using all activations from the full calibration set at once, we process a small number of full sequences at a time. Given the current rotations \(R_{1},\{R_{2,\ell}\}_{\ell=1}^{L}\), we run the model on a mini-batch of calibration sequences, extract activation vectors at the calibration sites of interest, normalize each activation vector, compute the targets induced by the current rotation, accumulate the corresponding Procrustes statistics, and then immediately update the rotation.


An important consequence of the online formulation is that it enables quantization-aware rotation optimization. Because quantizing activations at one layer alters the hidden states passed to subsequent layers, the activation distribution seen by later layers during inference differs from the full-precision distribution. The online procedure allows us to account for this shift by optimizing each rotation using the quantized activations produced by earlier layers. Accordingly, during calibration we quantize the input to each linear layer after applying the current rotation so that optimization is matched to inference-time conditions.

Algorithm~\ref{alg:online_calibration} summarizes the full online calibration procedure. For notational simplicity, the algorithm writes $R_{2,\ell}$ as a single matrix. In implementation, this matrix is
constrained to be block diagonal across attention heads, as described in
Appendix~\ref{rotation}. Thus, we solve a Procrustes problem per head. In addition, we do not need to store intermediate activations \(X_{\mathrm{attn}}^{(\ell)},X_{\mathrm{mlp}}^{(\ell)}\) across all transformer blocks as the cross covariance \(Z^{\top}\tilde{X}\) can be computed per layer and accumulated. This sequence-wise mini-batch scheme allows us to calibrate rotations efficiently without materializing all token activations from the entire calibration set, while still retaining the benefits of closed-form Procrustes updates.

\section{Experiments}
\label{sec:experiments}

\begin{table*}[t]
\centering
\scriptsize
\setlength{\tabcolsep}{3.75pt} 
\renewcommand{\arraystretch}{1.08}
\caption{Activation quantization results of rotation-based methods on Llama-2 and Llama-3 models. We report average perplexity across WikiText-2, PTB, and C4, and average zero-shot accuracy across nine common sense reasoning tasks.}
\label{tab:main_quant_table}
\begin{tabular}{c l *{5}{cc}}
\toprule
\multirow{2}{*}{\begin{tabular}{c}
Bits \\
(W-A-KV)
\end{tabular}}
& \multirow{2}{*}{Method}
& \multicolumn{2}{c}{Llama-2 7B}
& \multicolumn{2}{c}{Llama-2 13B}
& \multicolumn{2}{c}{Llama-3.2 3B}
& \multicolumn{2}{c}{Llama-3 8B}
& \multicolumn{2}{c}{Llama-3 70B} \\
\cmidrule(lr){3-4}
\cmidrule(lr){5-6}
\cmidrule(lr){7-8}
\cmidrule(lr){9-10}
\cmidrule(lr){11-12}
& 
& PPL$\downarrow$ & 0-shot$^9\uparrow$
& PPL$\downarrow$ & 0-shot$^9\uparrow$
& PPL$\downarrow$ & 0-shot$^9\uparrow$
& PPL$\downarrow$ & 0-shot$^9\uparrow$
& PPL$\downarrow$ & 0-shot$^9\uparrow$ \\
\midrule

16-16-16
& FloatingPoint
& 16.88 & 61.16
& 20.85 & 64.28
& 10.9 & 61.43
& 8.92 & 66.04
& 6.19 & 72.70 \\
\midrule

\multirow{6}{*}{4-4-16}
& QuaRot
& 20.34 & 57.89
& 24.82 & 62.36
& 13.75 & 55.64
& 11.06 & 61.12
& 13.26 & 57.53  \\
& DFRot
& \textbf{18.66} & 57.93
& 23.50 & \textbf{62.74}
& 13.76 & 55.66
& 10.96 & 61.52
& 11.70 & 60.72 \\
& SpinQuant
& 19.97 & 58.16
& 23.75 & \underline{62.50}
& 13.65 & 56.20
& 11.13 & 61.80
& 9.31 & 65.43 \\
& DartQuant
& 18.88 & \underline{58.25}
& 24.34 & 61.84
& 13.37 & \underline{56.75}
& 10.79 & \underline{61.96}
& 8.31 & \underline{68.16} \\
& \cellcolor{gray!20}Ours f.p. calib.
& \cellcolor{gray!20}19.28 & \cellcolor{gray!20}58.13
& \cellcolor{gray!20}\textbf{22.78} & \cellcolor{gray!20}62.13
& \cellcolor{gray!20}\underline{13.29} & \cellcolor{gray!20}56.13
& \cellcolor{gray!20}\underline{10.70} & \cellcolor{gray!20}61.80
& \cellcolor{gray!20}\textbf{8.25} & \cellcolor{gray!20}\textbf{68.17} \\
& \cellcolor{gray!20}Ours quant. calib.
& \cellcolor{gray!20}\underline{18.77} & \cellcolor{gray!20}\textbf{58.28}
& \cellcolor{gray!20}\underline{23.24} & \cellcolor{gray!20}62.43
& \cellcolor{gray!20}\textbf{13.21} & \cellcolor{gray!20}\textbf{56.79}
& \cellcolor{gray!20}\textbf{10.62} & \cellcolor{gray!20}\textbf{62.51}
& \cellcolor{gray!20}\underline{8.29} & \cellcolor{gray!20}67.63 \\
\midrule

\multirow{6}{*}{4-4-4}
& QuaRot
& 21.08 & 57.45
& 25.17 & 62.16
& 14.23 & 55.75
& 11.31 & 60.82
& 13.53 & 57.59 \\
& DFRot
& 19.18 & 57.78
& 23.57 & \textbf{62.26}
& 14.24 & 54.74
& 11.19 & 60.89
& 12.02 & 59.59 \\
& SpinQuant
& 20.69 & 57.62
& 24.05 & 61.62
& 13.98 & 55.43
& 11.28 & 60.95
& 9.06 & 65.52 \\
& DartQuant
& 19.57 & \textbf{58.02}
& 24.57 & 61.87
& 13.86 & \textbf{55.79}
& 11.00 & 61.20
& \underline{8.46} & 67.58 \\
& \cellcolor{gray!20}Ours f.p. calib.
& \cellcolor{gray!20}\underline{19.15} & \cellcolor{gray!20}\underline{57.99}
& \cellcolor{gray!20}\textbf{23.02} & \cellcolor{gray!20}61.89
& \cellcolor{gray!20}\underline{13.74} & \cellcolor{gray!20}54.87
& \cellcolor{gray!20}\underline{10.88} &  \cellcolor{gray!20}\underline{61.40}
& \cellcolor{gray!20}\textbf{8.40} & \cellcolor{gray!20}\underline{67.79} \\
& \cellcolor{gray!20}Ours quant. calib.
& \cellcolor{gray!20}\textbf{18.64} & \cellcolor{gray!20}57.89
& \cellcolor{gray!20}\underline{23.12} & \cellcolor{gray!20}\underline{62.24}
& \cellcolor{gray!20}\textbf{13.68} & \cellcolor{gray!20}\underline{55.77}
& \cellcolor{gray!20}\textbf{10.81} & \cellcolor{gray!20}\textbf{61.67}
& \cellcolor{gray!20}8.49 & \cellcolor{gray!20}\textbf{68.13} \\
\bottomrule
\end{tabular}
\end{table*}

We evaluate our method on the Llama-2~\cite{llama_2} and Llama-3~\cite{llama_3} model families, with model sizes ranging from 3B to 70B parameters. We compare against recent rotation-based methods for activation quantization, including QuaRot~\cite{quarot}, SpinQuant~\cite{spinquant}, DartQuant~\cite{dartquant}, and DFRot~\cite{dfrot}. We sample 128 calibration sequences of length 2,048 from WikiText-2~\cite{wikitext} and use this both for rotation calibration and weight quantization with GPTQ~\cite{optq}. Activations are quantized using asymmetric per-token quantization, and, when applicable, the KV cache is quantized using asymmetric quantization with group size 128. Additional implementation details are provided in Appendix~\ref{sec:implementation_details}.

We report perplexity (PPL) on WikiText-2~\cite{wikitext}, Penn Treebank (PTB)~\cite{ptb}, and C4~\cite{c4}. We also evaluate common-sense reasoning performance on nine downstream benchmarks: WinoGrande~\cite{winogrande}, SocialIQA~\cite{siqa}, LAMBADA~\cite{lambada}, MMLU~\cite{mmlu}, ARC-Easy, ARC-Challenge~\cite{arc}, HellaSwag~\cite{hellaswag}, OpenBookQA~\cite{obqa}, and PIQA~\cite{piqa}.

\subsection{Main results}

 We evaluate two quantization settings: 4-4-16, where weights and activations are quantized to 4 bits while the KV cache remains at 16 bits, and 4-4-4, where the KV cache is also quantized to 4 bits and present results in Table~\ref{tab:main_quant_table}.

Across both 4-4-16 and 4-4-4 settings, our method achieves strong perplexity relative to prior rotation-based methods. The quantization-aware variant obtains the best average PPL for most Llama-3.2 3B and Llama-3 8B settings and remains competitive on Llama-2 7B/13B and Llama-3 70B. These gains are most consistent when calibration is performed with activation quantization enabled, supporting the value of matching calibration to inference-time activation distributions.

On nine zero-shot reasoning benchmarks, our method remains competitive with the strongest baselines and often improves the average score. Thus, the PPL improvements do not come from uniformly degraded downstream behavior. Importantly, these results are achieved without the end-to-end rotation training used by SpinQuant or the stored activation corpus required by offline activation-calibration methods.

\subsection{Participation ratio analysis}

\begin{figure}
    \centering
    \includegraphics[width=1.0\linewidth]{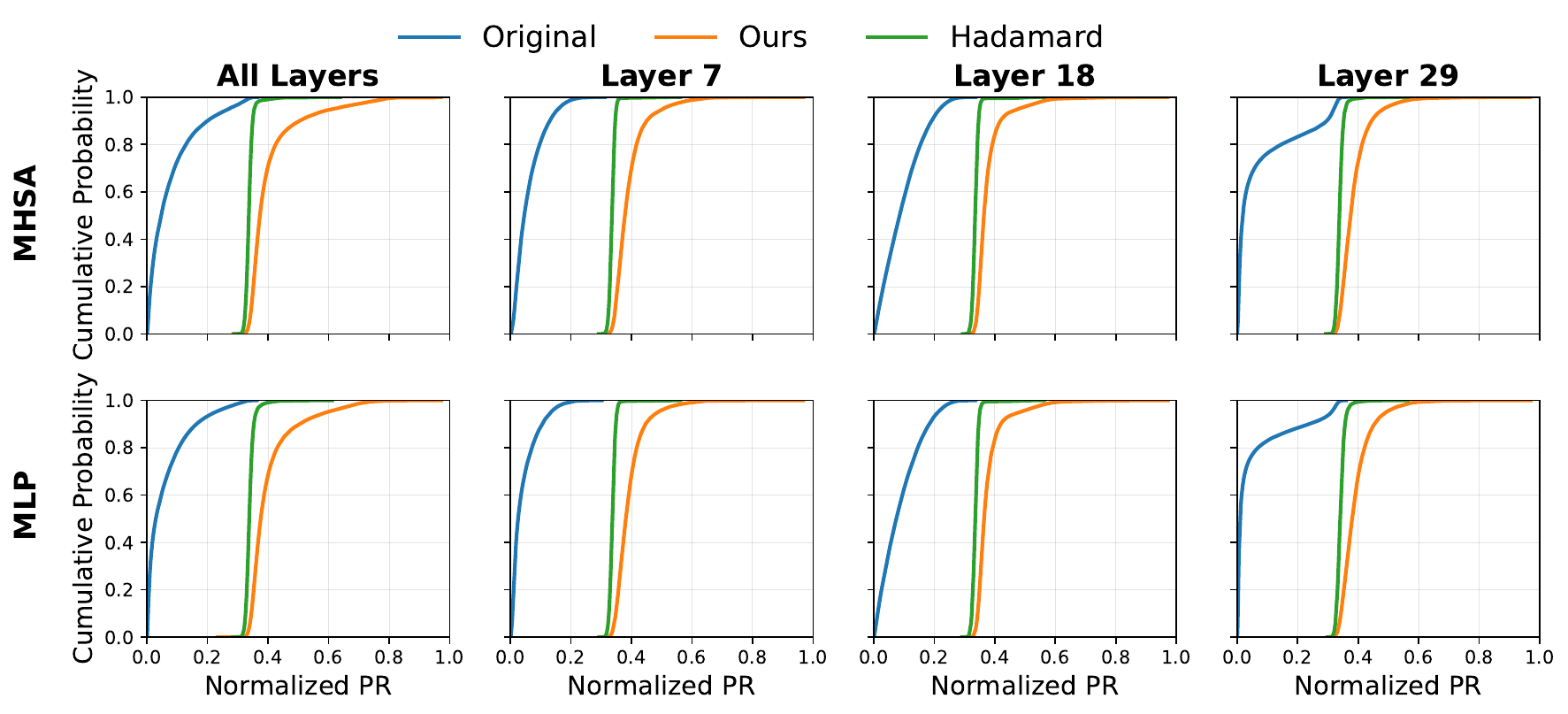}
    \caption{Empirical CDF of normalized participation ratio (PR) of activations from Llama-2 7B with different rotations.}  
    \label{fig:pr_analysis}
\end{figure}

The theory above suggests that effective rotations should move normalized
activations toward the equal-magnitude regime that minimizes
$\|R\tilde{x}\|_\infty$ on the unit sphere. To measure this behavior
empirically, we use the normalized participation ratio~\cite{pr1},
\[
\frac{\mathrm{PR}(y)}{d}
=
\frac{1}{d}\frac{\|y\|_2^4}{\|y\|_4^4}
\in [1/d,1].
\]
PR can be interpreted as the effective fraction of coordinates carrying
activation energy: it equals $1/d$ for a one-sparse vector and $1$ when all
coordinates have equal magnitude. Thus, larger normalized PR indicates that
activation energy is more evenly distributed across dimensions, the same
geometry targeted by our objective.

To evaluate this behavior empirically, we compare the normalized participation ratio, $\mathrm{PR}/d$, of activations from the calibration set under three settings: no rotation, Hadamard rotations, and our optimized rotations and present the results in Figure~\ref{fig:pr_analysis}.

Without rotation, the normalized PR is concentrated near zero, indicating that activation energy is highly concentrated in a small number of coordinates. Hadamard rotations substantially increase PR, with most activations attaining normalized PR values near $0.3$, suggesting that random mixing already spreads activation energy across a larger fraction of the feature dimensions. However, our optimized rotations consistently shift the distribution further to the right, yielding higher PR values across both MHSA and MLP activations. Moreover, our method shifts a noticeable fraction of activations into the higher-PR regime, with some activations reaching normalized PR values above $0.6$. Since normalized PR measures the effective fraction of dimensions over which activation energy is distributed, this indicates that our rotations substantially reduce feature concentration relative to the original activations and the Hadamard baseline.




\begin{table}
  \caption{Comparison between online calibration and an offline stored-activation variant on Llama-3 8B.}
  \label{tab:online_offline}
  \centering
  \begin{tabular}{clcccc}
    \toprule
    Bits (W/A/KV) & Method & WikiText2$\downarrow$      & PTB$\downarrow$ & C4$\downarrow$ & Avg$\downarrow$ \\
    \midrule
    \multirow{3}{*}{4-4-16} & Offline & 7.32 & 12.71 & 11.99 & 10.67\\
    & Online (f.p. calib.) & 7.35 & 12.81 & 11.93 & 10.70 \\
    & Online (quant. calib.) & \textbf{7.31} & \textbf{12.65} & \textbf{11.89} & \textbf{10.62}\\
    \midrule
    \multirow{3}{*}{4-4-4} & Offline & 7.45 & 12.95 & 12.23 & 10.88\\
    & Online (f.p. calib.) & 7.47 & 13.03 & 12.15 & 10.88 \\
    & Online (quant. calib.) & \textbf{7.44} & \textbf{12.89} & \textbf{12.11} & \textbf{10.81} \\
    \bottomrule
  \end{tabular}
\end{table}

\begin{figure}
    \centering
    \includegraphics[width=1.0\linewidth]{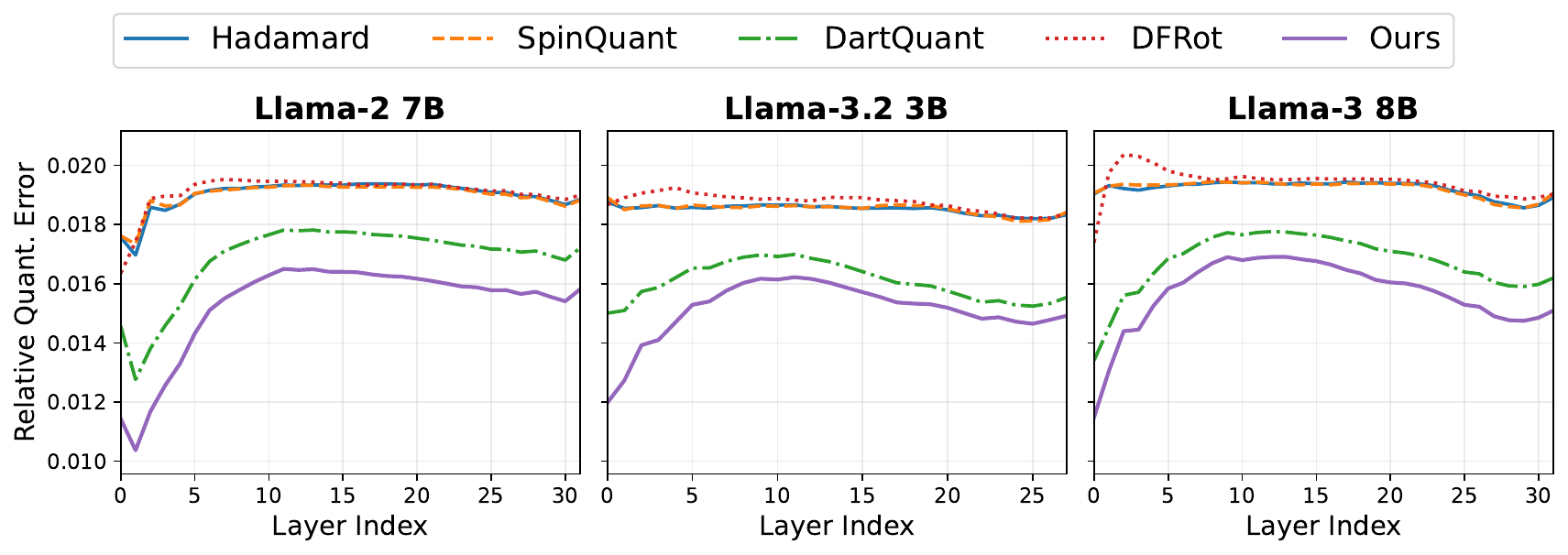}
    \caption{Layerwise activation quantization error for different rotation methods on Llama-2 7B, Llama-3.2 3B, and Llama-3 8B. We report the relative squared error between full-precision activations and their quantized reconstructions after applying each rotation.} 
    \label{fig:activation_quant_error}
\end{figure}

\subsection{Quantization error analysis}

To better understand the effect of the learned rotations, we directly measure activation quantization error across layers. For each model, we compare the relative squared quantization error 
\[
\frac{
\|X R^{\top} - Q(X R^{\top}) \|_{F}^{2}}{\| X \|_{F}^{2}}
\]
across rotations obtained by different methods. This analysis isolates the effect of the rotations on the activation distributions themselves, independent of downstream perplexity or task-level evaluation.

Figure~\ref{fig:activation_quant_error} shows these results for Llama-2 7B, Llama-3.2 3B, and Llama-3 8B. Across all three models, our method consistently yields the lowest relative error across all layers.  Compared with fixed Hadamard rotations and learned rotations from prior work, our method produces activations that are substantially easier to quantize, supporting the motivation of aligning normalized activations with hypercube corners to more evenly distribute activation magnitude across feature dimensions.

We also observe that DartQuant reduces quantization error relative to Hadamard, SpinQuant, and DFRot, but remains consistently above our method. This suggests that directly calibrating rotations toward quantization-friendly activation geometry provides a stronger reduction in quantization error than prior rotation objectives. These results help explain the perplexity improvements observed in Table~\ref{tab:main_quant_table}: by reducing the per-layer distortion introduced by activation quantization, our rotations preserve the full-precision computation more effectively under low-bit inference.

\subsection{Effect of online calibration}

A key component of our method is its online optimization. Rather than first storing a large corpus of intermediate activations and then optimizing rotations from this fixed dataset, our method updates the rotations directly as calibration samples are processed. This avoids the storage overhead of offline activation collection and also enables calibration under the same quantized activation distributions encountered at inference time.

To isolate the effect of this design choice, we compare our online procedure with an offline variant that stores all calibration activations to disk and then samples them i.i.d. during optimization. To ensure a fair comparison, we fix the number of activations used per Procrustes update and the number of activation steps. We evaluate both variants on Llama-3 8B and present results in Table~\ref{tab:online_offline}. The offline variant performs comparably to our online method when calibration is performed in full precision, indicating that the online updates closely approximate the behavior of optimization over stored i.i.d. activations. This suggests that our online formulation preserves the benefits of activation-based rotation optimization while avoiding the need to materialize the activation corpus. Lastly, we observe that enabling quantization during calibration improves performance supporting the motivation for quantization-aware calibration.


\begin{figure}
    \centering
    \includegraphics[width=0.95\linewidth]{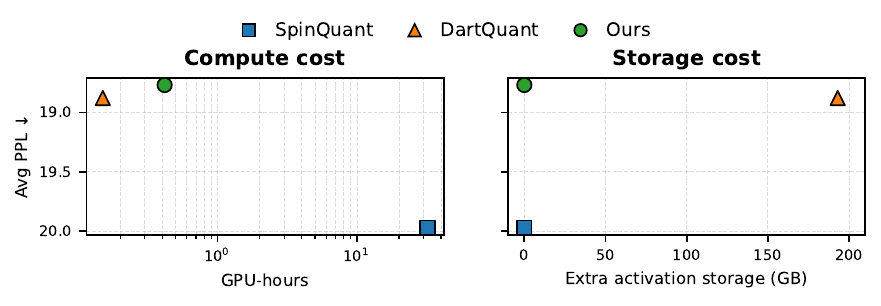}
    \caption{Calibration cost versus performance on Llama-2 7B under 4-4-16 quantization. Left: average PPL versus GPU-hours. Right: average PPL versus extra activation storage. Y-axis is inverted so upper-left is best.}
    \label{fig:calibration_tradeoff}
\end{figure}


\paragraph{Calibration efficiency.}
Figure~\ref{fig:calibration_tradeoff} compares the calibration cost--performance tradeoff for Llama-2 7B under 4-4-16 quantization. SpinQuant learns rotations through gradient-based optimization and, in our setup, requires 8 hours on 4 RTX A6000 GPUs (32 GPU-hours). DartQuant reduces calibration time to 9 minutes on a single RTX A6000, but does so by storing approximately 193GB of intermediate activations. Our method instead performs closed-form online calibration, requiring 25 minutes on a single RTX A6000 (0.42 GPU-hours) and no offline activation storage. Thus, our method occupies a favorable point in the tradeoff: it avoids SpinQuant's high compute cost and DartQuant's large storage overhead, while achieving the best average perplexity among these methods.

\section{Conclusion and limitations}
We introduced an efficient rotation-based calibration method for low-bit LLM activation quantization. By aligning normalized activations with the vertices of an inscribed hypercube, our method encourages activation energy to be distributed more evenly across feature dimensions, reducing the effect of outlier coordinates that dominate asymmetric quantization error. Our work avoids specialized optimizers for gradient-based optimization on the orthogonal group while also avoiding the computational and storage costs of prior works. Experiments across multiple LLM families and benchmarks show that this simple calibration procedure provides competitive quantized performance while substantially reducing calibration cost and storage overhead. These results suggest that closed-form, online rotation calibration is a practical direction for scalable post-training quantization of large language models. By reducing costs associated with rotation calibration, we improve the accessibility and ease of use of LLMs. However, if these models are misused, this could have negative societal impact.

\paragraph{Limitations} 
Our method adds a calibration step and requires optimized over a small set of calibration data, so performance may depend on the calibration data distribution. In addition, our experiments focus primarily on perplexity and common downstream benchmarks; broader evaluation of generation quality and latency remains important future work.

\section*{Acknowledgements}
This work was supported in part by the National Center on Generative AI for Uplifting STEM+C Education (GENIUS Center) and by the United States Air Force under Contract No. FA8750-23-C-0518. The authors also gratefully acknowledge the computing resources provided by the NVIDIA Academic Grant Program.

\newpage
\bibliographystyle{unsrtnat}
\bibliography{waq}

\newpage

\appendix

\section*{Appendix}

\section{Rotation invariance in LLMs}
\label{rotation}

Let $\hat{X} = XR$, where $X \in \mathbb{R}^{T \times d}$ denotes the activations of $T$ tokens and $R \in \mathbb{R}^{d \times d}$ is an orthogonal matrix. If $\hat{X}$ is passed to a linear layer with weight matrix $W \in \mathbb{R}^{d_o \times d}$, then the original computation can be preserved by absorbing the inverse rotation into the weight matrix:
\[
XW^\top = \hat{X}(R^\top W^\top).
\]
Thus, rotating the activations is equivalent to replacing $W$ by a rotated weight matrix, so no additional floating-point operations are required at inference for these mergeable rotations.

In pre-norm transformers, the residual stream is typically normalized before entering the self-attention and MLP blocks. For RMSNorm \cite{rmsnorm}, this does not break rotational invariance, since the normalization depends only on the vector norm. In practice, the learned RMSNorm scale is first fused into the adjacent linear weights, after which the rotation can be absorbed exactly into the surrounding weight matrices. This is the same computational-invariance mechanism used by prior rotation-based quantization methods.

Following prior work \cite{quarot, spinquant, dartquant, dfrot}, we optimize a single rotation shared across all layers denoted as $R_1$ as well as a layer specific rotation denoted as $R_{2,\ell}$. The rotation $R_1$ acts on the residual stream and is shared across all transformer blocks. It is applied to the inputs of the self-attention and MLP sublayers, and its inverse is absorbed into the corresponding input projections. Because the residual stream is propagated through skip connections, the same $R_1$ must be used throughout the network; correspondingly, the output projections that write back to the residual stream absorb the matching inverse transform. In our notation, this means that $R_1$ is merged into the weights associated with the attention and MLP inputs, while the inverse transform is absorbed into the output projections $W_O$ and $W_{\mathrm{down}}$.

We additionally use a layer-specific rotation $R_{2,\ell}$ inside each attention block. This rotation is applied to the activations input to the output projection $W_O$ and can therefore be paired with a corresponding transform on the value path. Concretely, $R_{2,\ell}$ is absorbed into the value projection $W_V$, while its inverse is absorbed into $W_O$, yielding an equivalent full-precision computation but a more quantization-friendly activation distribution at the input to the output projection. As these rotations do not need to be shared across layers, each $R_{2,\ell}$ rotation is learned separately for each transformer block. Lastly, due to the self-attention operation being performed across multiple heads, $R_{2,\ell}$ is constrained to be block diagonal with block size \(d_{h}\times d_{h}\).

The rotations above are the only ones we explicitly optimize and merge into the model weights. In addition, following QuaRot\cite{quarot} and SpinQuant\cite{spinquant}, we optionally use online Hadamard transforms to further reduce outliers in locations where mergeable rotations are not sufficient or cannot be absorbed as conveniently. We denote these fixed transforms by $R_3$ and $R_4$. The transform $R_3$ is applied after the rotary positional embeddings (RoPE)\cite{rope} within self-attention to improve KV-cache quantization, while $R_4$ is applied before the down projection to improve low-bit activation quantization inside the MLP. Because Hadamard transforms admit efficient fast implementations, they can be applied online during inference with modest overhead. 

\section{From quantization error to the corner objective derivation}
\label{app:quant_error}

We now provide a theoretical justification for aligning activations with the
corners of the inscribed hypercube. The argument proceeds in three steps:
(i) the mean-squared error of both uniform symmetric and asymmetric (zeropoint) quantizers is controlled by
the ratio $\|x\|_\infty^2 / \|x\|_2^2$; (ii) on the unit sphere, this ratio
attains its minimum exactly at the hypercube corners; (iii) the
corner-distance objective in equation~\ref{eq:objective} is equivalent to a
smooth $\ell_1$-maximization that serves as a tight, differentiable proxy for
minimizing $\|x\|_\infty$.

\paragraph{Quantization error of a per-tensor symmetric uniform quantizer.}
Consider a symmetric $b$-bit uniform quantizer applied to an activation
vector $x \in \mathbb{R}^d$ with per-tensor step size determined by the
absolute maximum coordinate,
\[
Q_{\mathrm{sym}}(x) \;=\; \Delta_{\mathrm{sym}}(x)\,\mathrm{round}\!\left(\frac{x}{\Delta_{\mathrm{sym}}(x)}\right),
\qquad
\Delta_{\mathrm{sym}}(x) \;=\; \frac{\|x\|_\infty}{2^{b-1}-1}.
\]
Each coordinate incurs a rounding error bounded in magnitude by
$\Delta_{\mathrm{sym}}(x)/2$, so the total mean-squared quantization error
satisfies
\begin{equation}\label{eq:mse_bound_sym}
\|Q_{\mathrm{sym}}(x)-x\|_2^2 \;\le\; \frac{d\,\Delta_{\mathrm{sym}}(x)^2}{4}
\;=\; \frac{d}{4(2^{b-1}-1)^2} \, \|x\|_\infty^2.
\end{equation}
The signal-to-quantization-noise ratio (SQNR) is therefore lower bounded by
\begin{equation}\label{eq:sqnr_bound}
\mathrm{SQNR}(x) \;=\; \frac{\|x\|_2^2}{\|Q_{\mathrm{sym}}(x)-x\|_2^2}
\;\ge\; \frac{4(2^{b-1}-1)^2}{d}\cdot\frac{1}{\rho(x)},
\qquad
\rho(x) \;:=\; \frac{\|x\|_\infty^2}{\|x\|_2^2}.
\end{equation}
The dynamic-range ratio $\rho(x) \in [1/d, 1]$ is the only data-dependent
quantity in the bound: minimizing $\rho$ over the unit sphere directly
maximizes the worst-case SQNR.

\paragraph{Quantization error of a per-tensor asymmetric (zeropoint) quantizer.}
A zeropoint quantizer \cite{asym_quant} maps the affine range
$[\min_j x_j,\,\max_j x_j]$ onto the full unsigned integer grid
$\{0,\ldots,2^{b}-1\}$ via a scale and a shift,
\[
Q_{\mathrm{zp}}(x) \;=\; \Delta_{\mathrm{zp}}(x)\,
\Bigl(\mathrm{round}\!\left(\tfrac{x}{\Delta_{\mathrm{zp}}(x)}+\zeta(x)\right) - \zeta(x)\Bigr),
\quad
\Delta_{\mathrm{zp}}(x) \;=\; \frac{\max_j x_j - \min_j x_j}{2^{b}-1},
\]
where $\zeta(x) \in \mathbb{Z}$ is the integer zeropoint that aligns
$\min_j x_j$ with the lower end of the grid. Each coordinate again incurs a
rounding error bounded by $\Delta_{\mathrm{zp}}(x)/2$, yielding
\begin{equation}\label{eq:mse_bound_zp}
\|Q_{\mathrm{zp}}(x)-x\|_2^2 \;\le\; \frac{d\,\Delta_{\mathrm{zp}}(x)^2}{4}
\;=\; \frac{d}{4(2^{b}-1)^2}\,\bigl(\max_j x_j - \min_j x_j\bigr)^2.
\end{equation}
The data-dependent quantity is now the signed range
$\max_j x_j - \min_j x_j$ rather than $\|x\|_\infty$. These two are tied by
the elementary inequality
\begin{equation}\label{eq:range_linfty}
\max_j x_j - \min_j x_j \;\le\; 2\|x\|_\infty,
\end{equation}
with equality whenever $x$ has at least one coordinate of each sign achieving
the maximum absolute value---in particular at every sign-mixed vertex of the
inscribed hypercube. Combining \eqref{eq:mse_bound_zp} and
\eqref{eq:range_linfty} gives
\begin{equation}\label{eq:mse_bound_zp_via_linfty}
\|Q_{\mathrm{zp}}(x)-x\|_2^2 \;\le\; \frac{d}{(2^{b}-1)^2}\,\|x\|_\infty^2,
\end{equation}
matching the symmetric bound \eqref{eq:mse_bound_sym}.

\paragraph{Tightness on the unit sphere.}
A keen reader may worry that \eqref{eq:range_linfty} can be loose: the bound
upper-bounds the signed range by $2\|x\|_\infty$, but the two quantities
need not share the same minimizer on the unit sphere. We address this
concern by quantifying both sides directly. For $x \in \mathbb{S}^{d-1}$,
\[
0 \;\le\; \max_j x_j - \min_j x_j \;\le\; \sqrt{2},
\qquad
\frac{1}{\sqrt{d}} \;\le\; \|x\|_\infty \;\le\; 1.
\]
The signed range attains its minimum of $0$ at the all-same-sign directions
$\pm \mathbf{1}/\sqrt{d}$, while $\|x\|_\infty$ attains its minimum
$1/\sqrt{d}$ at every vertex of the inscribed hypercube. The two surrogates
therefore disagree on the global optimum: minimizing $\|x\|_\infty$ drives
$x$ to an arbitrary hypercube corner, whereas minimizing the signed range
would prefer the two specific same-sign corners.

The disagreement, however, vanishes with dimension. At any sign-mixed
hypercube corner---the generic fixed point of the corner objective---the
signed range equals $2/\sqrt{d}$, while the unreachable global minimum is
$0$. The absolute gap is $O(1/\sqrt{d})$, and the achieved value lies
within a fraction $\sqrt{2/d}$ of the worst-case range $\sqrt{2}$. For
transformer-scale hidden dimensions this gap is negligible: at $d=4096$,
sign-mixed corners achieve a range of $2/\sqrt{4096} \approx 0.031$,
recovering more than $97\%$ of the available range reduction on the unit
sphere. The factor-of-two slack in \eqref{eq:range_linfty} is loose at
axis poles---which the optimization moves away from---and asymptotically
tight at the corners it converges to. Minimizing $\|x\|_\infty$ therefore
controls the MSE bound for both quantizers simultaneously, with a
zeropoint-side suboptimality that decays as $1/\sqrt{d}$ and is
quantitatively immaterial in the regime of interest. The remainder of this
section may be read with $\|x\|_\infty$ as the single quantity to
minimize, with the understanding that the analysis applies verbatim to
symmetric quantization and, up to vanishing dimensional corrections, to
asymmetric quantization as well.

\paragraph{Corners minimize the dynamic-range ratio.}
An orthogonal rotation $R \in O(d)$ preserves $\ell_2$ norms, so
$\|R\tilde{x}\|_2 = 1$ for any normalized activation $\tilde{x}$. Rotation
therefore operates entirely through the numerator of $\rho$:
\[
\rho(R\tilde{x}) \;=\; \|R\tilde{x}\|_\infty^2.
\]
The following standard fact identifies the minimizers.

\begin{proposition}\label{prop:corners}
For any $y \in \mathbb{R}^d$ with $\|y\|_2=1$, $\|y\|_\infty^2 \ge 1/d$, with
equality if and only if $y$ is a vertex of the hypercube inscribed in the
unit sphere, i.e.\ $y = \frac{1}{\sqrt{d}}(\pm 1,\ldots,\pm 1)$.
\end{proposition}

Combined with~\eqref{eq:sqnr_bound}, this shows that hypercube corners are the directions on the sphere at which the quantization error bound
attains its minimum value

\[
\|Q_{\mathrm{sym}}(y)-y\|_2^2
\le
\frac{1}{4(2^{b-1}-1)^2}.
\]

For comparison, axis
poles $\pm e_j$ achieve the worst case $\rho = 1$, in which case
the bound reduces to 

\[
\|Q_{\mathrm{sym}}(y)-y\|_2^2
\le
\frac{d}{4(2^{b-1}-1)^2}.
\]

which is $d$ larger than the corner regime.

\paragraph{Dualization: an $\ell_1$ surrogate for $\ell_\infty$.}
Although Proposition~\ref{prop:corners} identifies the ideal target,
$\|y\|_\infty$ is non-smooth and its subgradient concentrates on a single
coordinate, making direct minimization over $O(d)$ ill-conditioned. We
instead work with the dual norm via its variational representation. Recall that
$\ell_1$ and $\ell_\infty$ are Hölder conjugates, with
\begin{equation}\label{eq:dual_repr}
\|y\|_1 \;=\; \max_{s \in \{-1,+1\}^d} \langle s,\, y\rangle
\;=\; \langle \mathrm{sign}(y),\, y\rangle.
\end{equation}
On the unit sphere, Cauchy--Schwarz gives
\begin{equation}\label{eq:cs_bound}
\|y\|_1 \;\le\; \sqrt{d}\,\|y\|_2 \;=\; \sqrt{d},
\end{equation}
with equality if and only if $|y_1|=\cdots=|y_d|=1/\sqrt{d}$, i.e.,
$y$ is a hypercube corner. Thus, maximizing $\|R\tilde{x}\|_1$ has the
same set of per-sample maximizers on the unit sphere as minimizing
$\|R\tilde{x}\|_\infty$, while providing dense subgradient information
across coordinates. Substituting~\eqref{eq:dual_repr} introduces auxiliary
sign variables $s^i \in \{-1,+1\}^d$ and yields the equivalent problem
\begin{equation}\label{eq:dualized}
\max_{R \in O(d)} \;\sum_i \max_{s^i \in \{-1,+1\}^d}
\langle s^i,\, R\tilde{x}^i\rangle.
\end{equation}
Fixing the sign variables makes the objective linear in $R$, and the
resulting maximization over $O(d)$ reduces to an orthogonal Procrustes
problem. This alternating maximization—updating
$s^i \leftarrow \mathrm{sign}(R\tilde{x}^i)$ followed by a closed-form
Procrustes update for $R$—is exactly the procedure used in
Section~\ref{sec:procrustes}.

\paragraph{Equivalence to the corner-distance objective.}
The corner-distance objective in Equation~\ref{eq:objective} is the
rescaled form of the dualized problem~\eqref{eq:dualized} in disguise: identifying
$z^i = s^i/\sqrt{d}$ recovers the targets defined coordinatewise as
$z^i_j = \tfrac{1}{\sqrt{d}}\,\mathrm{sign}((R\tilde{x}^i)_j)$. To see this
explicitly, expand the squared distance:
\begin{align*}
\|R\tilde{x}^i - z^i\|_2^2
&= \|R\tilde{x}^i\|_2^2 + \|z^i\|_2^2 - 2\,\langle R\tilde{x}^i,\,z^i\rangle \\
&= 2 \;-\; \frac{2}{\sqrt{d}}\sum_{j=1}^d (R\tilde{x}^i)_j\,\mathrm{sign}((R\tilde{x}^i)_j) \\
&= 2 \;-\; \frac{2}{\sqrt{d}}\,\|R\tilde{x}^i\|_1,
\end{align*}
where the last equality uses the variational identity~\eqref{eq:dual_repr}
evaluated at the optimal sign vector $s^i = \mathrm{sign}(R\tilde{x}^i)$.
Summing over the calibration set yields
\begin{equation}\label{eq:corner_l1}
\sum_i \|R\tilde{x}^i - z^i\|_2^2
\;=\; 2N \;-\; \frac{2}{\sqrt{d}}\sum_i \|R\tilde{x}^i\|_1,
\end{equation}
so that minimizing the corner distance is exactly equivalent to maximizing
the average $\ell_1$ norm of the rotated, normalized activations—the outer
problem of~\eqref{eq:dualized} after the inner sign maximization has been
solved in closed form. By the Cauchy--Schwarz bound~\eqref{eq:cs_bound}
and Proposition~\ref{prop:corners}, this is a tight proxy for jointly
minimizing $\|R\tilde{x}^i\|_\infty$ for each $i$, and hence for
minimizing the upper bound~\eqref{eq:mse_bound_zp} on quantization MSE in
expectation across the calibration distribution.


\section{Implementation details}
\label{sec:implementation_details}

\paragraph{Implementation details.}
We generate optimized rotations using the publicly available implementations of
SpinQuant~\cite{spinquant}, DFRot~\cite{dfrot}, and DartQuant~\cite{dartquant}. For Llama-3 70B, we use the publicly available rotations for SpinQuant which are trained on a larger corpus of calibration data. 
All methods are evaluated with a common evaluation pipeline adapted from the
public DartQuant codebase. Unless otherwise stated, calibration is performed on
the same 128 WikiText2~\cite{wikitext} samples for all methods to ensure a fair
comparison. Since DFRot optimizes its rotation using a single calibration sample,
we use the first sample from this shared calibration set.

For weight quantization, quantization scales are computed per output row. For
activation and KV-cache quantization, we follow the DartQuant evaluation setting
and use clipping ratios of \(0.9\) and \(1.0\), respectively. To evaluate QuaRot~\cite{quarot} within our rotation framework, we replace the
learned rotations \(R_1\) and \(R_{2,\ell}\) with random Hadamard rotations,
implemented using the Quip\# Hadamard routines~\cite{quip_sharp}.
Our calibration procedure uses batch size 1 for all models, except for
Llama-2 7B and Llama-3 8B, where we use batch size 2.

Experiments on Llama-2 7B, 13B, Llama-3.2 3B, and Llama-3 8B are all completed on servers with 4 RTX A6000 GPUs with calibration performed on a single GPU unless otherwise stated. For SpinQuant, we use all 4 GPUs during calibration. Llama-3 70B experiments are all completed using a single RTX Pro 6000 Blackwell GPU.

\section{Full results}

Below in Tables \ref{tab:ppl_2_7b} to \ref{tab:common_sense_3_70b} we present the full results used to populate Table \ref{tab:main_quant_table}. We present perplexity results on WikiText2~\cite{wikitext}, Penn TreeBank (PTB)~\cite{ptb}, and C4~\cite{c4}. For common sense reasoning, we evaluate on WinoGrande (WG)~\cite{winogrande}, SocialIQA (SIQA)~\cite{siqa}, LAMBADA (LAMB)~\cite{lambada}, MMLU~\cite{mmlu}, ARC-Easy (ARC-E), ARC-Challenge (ARC-C)~\cite{arc}, HellaSwag (HS)~\cite{hellaswag}, OpenBookQA (OBQA)~\cite{obqa}, and PIQA~\cite{piqa}.

\begin{table}
  \caption{Full perplexity results for Llama-2 7B}
  \label{tab:ppl_2_7b}
  \centering
  \begin{tabular}{clcccc}
    \toprule
    Bits (W/A/KV) & Method & WikiText2$\downarrow$      & PTB$\downarrow$ & C4$\downarrow$ & Avg$\downarrow$ \\
    \midrule
    \multirow{6}{*}{4-4-16} & QuaRot & 6.04 & 46.79 & 8.18 & 20.34 \\
    & DFRot & 5.97 & 41.91 & 8.10 & 18.66 \\
    & SpinQuant & 6.02 & 45.72 & 8.16 & 19.97 \\
    & DartQuant & 5.96 & 42.62 & 8.07 & 18.88 \\
    & Ours f.p. calib. & 5.94 & 43.86 & 8.04 & 19.28 \\
    & Ours quant. calib. & 5.93 & 42.34 & 8.04 & 18.77 \\
    \midrule
    \multirow{6}{*}{4-4-4} & QuaRot & 6.11 & 48.86 & 8.27 & 21.08 \\
    & DFRot & 6.03 & 43.33 & 8.18 & 19.18 \\
    & SpinQuant & 6.09 & 47.70 & 8.27 & 20.69 \\
    & DartQuant & 6.02 & 44.52 & 8.17 & 19.57 \\
    & Ours f.p. calib. & 6.00 & 43.33 & 8.13 & 19.15 \\
    & Ours quant. calib. & 6.00 & 41.74 & 8.17 & 18.64 \\
    \bottomrule
  \end{tabular}
\end{table}

\begin{table}
\centering
\scriptsize
\setlength{\tabcolsep}{5pt} 
\renewcommand{\arraystretch}{1.08}
  \caption{Full common sense reasoning results for Llama-2 7B}
  \label{tab:common_sense_2_7b}
  \centering
  \begin{tabular}{cl ccccccccc c}
    \toprule
    Bits (W/A/KV) & Method & WG      & SIQA & LAMB & MMLU & ARC-E & ARC-C & HS & OBQA & PIQA & Avg\\
    \midrule
    \multirow{6}{*}{4-4-16} & QuaRot & 67.09 & 43.55 & 72.04 & 34.6 & 70.58 & 42.66 & 73.31 & 39.6 & 77.58 & 57.89 \\
    & DFRot & 67.17 & 44.11 & 71.96 & 34.75 & 69.81 & 42.41 & 73.0 & 40.8 & 77.37 & 57.93 \\
    & SpinQuant & 66.93  & 43.35 & 72.04 & 34.70 & 71.80 & 42.41 & 73.09 & 41.60 & 77.48 & 58.16 \\
    & DartQuant & 65.11 & 44.06 & 70.62 & 37.07 & 72.60 & 44.11 & 73.32 & 40.20 & 77.20 & 58.25 \\
    & Ours f.p. calib. & 66.22 & 44.52 & 71.76 & 33.21 & 71.09 & 43.0 & 73.68 & 41.8 & 77.86 & 58.13 \\
    & Ours quant. calib. & 66.69 & 44.88 & 72.68 & 33.56 & 71.04 & 42.41 & 73.41 & 42.40 & 77.48 & 58.28 \\
    \midrule
    \multirow{6}{*}{4-4-4} & QuaRot & 66.61 & 43.86 & 70.77 & 34.44 & 69.49 & 41.81 & 72.86 & 40.2 & 77.04 & 57.45\\
    & DFRot & 66.61 & 44.58 & 71.45 & 34.55 & 69.40 & 42.06 & 72.73 & 42.2 & 76.33 & 57.78\\
    & SpinQuant & 65.59 & 44.17 & 71.1 & 35.07 & 70.54 & 40.53 & 72.79 & 40.8 & 77.97 & 57.62\\
    & DartQuant & 65.11 & 42.84 & 71.01 & 36.3 & 71.46 & 43.26 & 73.41 & 41.0 & 77.8 & 58.02\\
    & Ours f.p. calib. & 67.48 & 44.42 & 71.63 & 34.07 & 70.37 & 41.13 & 73.23 & 42.0 & 77.58 & 57.99\\
    & Ours quant. calib. & 67.88 & 44.11 & 70.79 & 34.33 & 70.33 & 42.75 & 72.81 & 40.4 & 77.64 & 57.89\\
    \bottomrule
  \end{tabular}
\end{table}

\begin{table}
  \caption{Full perplexity results for Llama-2 13B}
  \label{tab:ppl_2_13b}
  \centering
  \begin{tabular}{clcccc}
    \toprule
    Bits (W/A/KV) & Method & WikiText2$\downarrow$      & PTB$\downarrow$ & C4$\downarrow$ & Avg$\downarrow$ \\
    \midrule
    \multirow{6}{*}{4-4-16} & QuaRot & 5.29 & 61.78 & 7.38 & 24.82 \\
    & DFRot & 5.25 & 57.93 & 7.31 & 23.50 \\
    & SpinQuant & 5.32 & 58.58 & 7.36 & 23.75 \\
    & DartQuant & 5.29 & 60.35 & 7.37 & 24.34\\
    & Ours f.p. calib. & 5.25 & 55.78 & 7.31 & 22.78 \\
    & Ours quant. calib. & 5.25 & 57.13 & 7.33 & 23.24 \\
    \midrule
    \multirow{6}{*}{4-4-4} & QuaRot & 5.33 & 62.76 & 7.43 & 25.17  \\
    & DFRot & 5.29 & 58.06 & 7.36 & 23.57 \\
    & SpinQuant & 5.37 & 59.34 & 7.44 & 24.05 \\
    & DartQuant & 5.34 & 60.95 & 7.43 & 24.57  \\
    & Ours f.p. calib. & 5.29 & 56.41 & 7.37 & 23.02\\
    & Ours quant. calib. & 5.30 & 57.67 & 7.40 & 23.12 \\
    \bottomrule
  \end{tabular}
\end{table}

\begin{table}
\centering
\scriptsize
\setlength{\tabcolsep}{5pt} 
\renewcommand{\arraystretch}{1.08}
  \caption{Full common sense reasoning results for Llama-2 13B}
  \label{tab:common_sense_2_13b}
  \centering
  \begin{tabular}{cl ccccccccc c}
    \toprule
    Bits (W/A/KV) & Method & WG      & SIQA & LAMB & MMLU & ARC-E & ARC-C & HS & OBQA & PIQA & Avg\\
    \midrule
    \multirow{6}{*}{4-4-16} & QuaRot & 68.51 & 45.7 & 75.78 & 47.08 & 74.92 & 48.12 & 76.79 & 44.6 & 79.76 & 62.36 \\
    & DFRot & 71.82 & 46.52 & 75.59 & 46.3 & 75.97 & 48.21 & 76.93 & 44.6 & 78.73 & 62.74 \\
    & SpinQuant & 71.27 & 46.01 & 74.85 & 47.71 & 75.21 & 47.61 & 76.26 & 44.2 & 79.38 & 62.5 \\
    & DartQuant & 71.51 & 43.71 & 75.41 & 44.19 & 74.07 & 47.7 & 77.19 & 43.8 & 78.94 & 61.84 \\
    & Ours f.p. calib. & 69.77 & 44.98 & 74.54 & 48.1 & 74.75 & 47.95 & 75.9 & 44.0 & 79.16 & 62.13 \\
    & Ours quant. calib. & 70.8 & 46.21 & 75.16 & 47.17 & 75.21 & 48.55 & 77.45 & 42.4 & 78.89 & 62.43 \\
    \midrule
    \multirow{6}{*}{4-4-4} & QuaRot & 69.14 & 45.55 & 75.33 & 46.08 & 75.04 & 48.81 & 76.78 & 43.4 & 79.33 & 62.16\\
    & DFRot & 69.93 & 45.8 & 75.02 & 45.86 & 75.88 & 48.46 & 77.26 & 43.2 & 78.94 & 62.26\\
    & SpinQuant & 69.38 & 45.04 & 74.44 & 46.22 & 74.66 & 46.59 & 76.52 & 43.4 & 78.29 & 61.62 \\
    & DartQuant & 69.93 & 44.47 & 74.85 & 47.03 & 74.28 & 45.82 & 76.75 & 44.8 & 78.89 & 61.87\\
    & Ours f.p. calib. & 71.27 & 44.27 & 74.73 & 47.33 & 74.37 & 46.93 & 75.64 & 43.4 & 79.05 & 61.89\\
    & Ours quant. calib. & 70.72 & 46.01 & 74.91 & 46.72 & 75.17 & 47.1 & 77.18 & 43.8 & 78.51 & 62.24\\
    \bottomrule
  \end{tabular}
\end{table}

\begin{table}
  \caption{Full perplexity results for Llama-3.2 3B}
  \label{tab:ppl_3_3b}
  \centering
  \begin{tabular}{clcccc}
    \toprule
    Bits (W/A/KV) & Method & WikiText2$\downarrow$      & PTB$\downarrow$ & C4$\downarrow$ & Avg$\downarrow$ \\
    \midrule
    \multirow{6}{*}{4-4-16} & QuaRot & 9.62 & 16.65 & 14.99 & 13.75   \\
    & DFRot & 9.65 & 16.63 & 15.01 & 13.76\\
    & SpinQuant & 9.51 & 16.58 & 14.86 & 13.65 \\
    & DartQuant & 9.35 & 16.16 & 14.59 & 13.37\\
    & Ours f.p. calib. & 9.25 & 16.19 & 14.43 & 13.29 \\
    & Ours quant. calib. & 9.19 & 16.10 & 14.33 & 13.21 \\
    \midrule
    \multirow{6}{*}{4-4-4} & QuaRot & 9.91 & 17.28 & 15.50 & 14.23  \\
    & DFRot & 9.94 & 17.30 & 15.48 & 14.24\\
    & SpinQuant & 9.73 & 16.98 & 15.24 & 13.98\\
    & DartQuant & 9.63 & 16.91 & 15.05 & 13.86\\
    & Ours f.p. calib. & 9.50 & 16.83 & 14.90 & 13.74\\
    & Ours quant. calib. & 9.49 & 16.72 & 14.84 & 13.68 \\
    \bottomrule
  \end{tabular}
\end{table}

\begin{table}
\centering
\scriptsize
\setlength{\tabcolsep}{5pt} 
\renewcommand{\arraystretch}{1.08}
  \caption{Full common sense reasoning results for Llama-3.2 3B}
  \label{tab:common_sense_3_3b}
  \centering
  \begin{tabular}{cl ccccccccc c}
    \toprule
    Bits (W/A/KV) & Method & WG      & SIQA & LAMB & MMLU & ARC-E & ARC-C & HS & OBQA & PIQA & Avg\\
    \midrule
    \multirow{6}{*}{4-4-16} & QuaRot & 62.12 & 43.86 & 62.7 & 46.2 & 64.6 & 38.91 & 68.93 & 39.0 & 74.48 & 55.64\\
    & DFRot & 62.51 & 44.73 & 61.13 & 45.7 & 63.89 & 41.04 & 68.42 & 40.0 & 73.5 & 55.66\\
    & SpinQuant & 63.93 & 44.27 & 63.3 & 46.54 & 66.41 & 40.36 & 68.48 & 38.0 & 74.54 & 56.20\\
    & DartQuant & 64.72 & 44.11 & 63.63 & 46.23 & 66.71 & 42.58 & 69.36 & 38.8 & 74.65 & 56.75\\
    & Ours f.p. calib. & 64.4 & 44.98 & 62.53 & 44.7 & 64.10 & 39.76 & 69.26 & 40.0 & 75.41 & 56.13 \\
    & Ours quant. calib. & 65.19 & 44.37 & 64.41 & 46.01 & 66.50 & 40.78 & 68.75 & 70.6 & 74.54 & 56.79 \\
    \midrule
    \multirow{6}{*}{4-4-4} & QuaRot & 64.09 & 45.04 & 61.5 & 44.06 & 64.77 & 40.19 & 68.0 & 40.0 & 74.10 & 55.75\\
    & DFRot & 63.3 & 44.32 & 59.27 & 44.19 & 62.79 & 38.4 & 68.17 & 37.60 & 74.59 & 54.74\\
    & SpinQuant & 63.14 & 43.45 & 62.10 & 44.75 & 64.44 & 40.36 & 68.16 & 37.40 & 75.03 & 55.43\\
    & DartQuant & 62.98 & 42.99 & 62.99 & 44.32 & 66.04 & 42.15 & 68.20 & 38.40 & 74.05 & 55.79\\
    & Ours f.p. calib. & 61.80 & 44.37 & 60.78 & 42.27 & 62.33 & 40.10 & 68.43 & 39.80 & 73.94 & 54.87\\
    & Ours quant. calib. & 64.25 & 44.78 & 62.31 & 44.75 & 63.09 & 39.85 & 68.72 & 39.60 & 74.59 & 55.77\\
    \bottomrule
  \end{tabular}
\end{table}

\begin{table}
  \caption{Full perplexity results for Llama-3 8B}
  \label{tab:ppl_3_8b}
  \centering
  \begin{tabular}{clcccc}
    \toprule
    Bits (W/A/KV) & Method & WikiText2$\downarrow$      & PTB$\downarrow$ & C4$\downarrow$ & Avg$\downarrow$ \\
    \midrule
    \multirow{6}{*}{4-4-16} & QuaRot & 7.62 & 13.21 & 12.34 & 11.06  \\
    & DFRot & 7.58 & 12.91 & 12.38 & 10.96  \\
    & SpinQuant & 7.64 & 13.39 & 12.37 & 11.13 \\
    & DartQuant & 7.42 & 12.92 & 12.04 & 10.79 \\
    & Ours f.p. calib. & 7.35 & 12.81 & 11.93 & 10.70 \\
    & Ours quant. calib. & 7.31 & 12.65 & 11.89 & 10.62 \\
    \midrule
    \multirow{6}{*}{4-4-4} & QuaRot & 7.78 & 13.53 & 12.62 & 11.31  \\
    & DFRot & 7.74 & 13.22 & 12.61 & 11.19 \\
    & SpinQuant & 7.75 & 13.52 & 12.58 & 11.28 \\
    & DartQuant & 7.54 & 13.19 & 12.28 & 11.00  \\
    & Ours f.p. calib. & 7.47 & 13.03 & 12.15 & 10.88\\
    & Ours quant. calib. & 7.44 & 12.89 & 12.11 & 10.81 \\
    \bottomrule
  \end{tabular}
\end{table}

\begin{table}
\centering
\scriptsize
\setlength{\tabcolsep}{5pt} 
\renewcommand{\arraystretch}{1.08}
  \caption{Full common sense reasoning results for Llama-3 8B}
  \label{tab:common_sense_3_8b}
  \centering
  \begin{tabular}{cl ccccccccc c}
    \toprule
    Bits (W/A/KV) & Method & WG      & SIQA & LAMB & MMLU & ARC-E & ARC-C & HS & OBQA & PIQA & Avg\\
    \midrule
    \multirow{6}{*}{4-4-16} & QuaRot & 66.85 & 44.52 & 69.88 & 54.59 & 74.03 & 46.33 & 75.28 & 41.60 & 76.99 & 61.12 \\
    & DFRot & 69.93 & 45.24 & 71.78 & 55.12 & 73.91 & 45.9 & 75.54 & 40.0 & 76.28 & 61.52 \\
    & SpinQuant & 68.19 & 44.47 & 69.75 & 54.93 & 76.01 & 48.21 & 75.62 & 41.40 & 77.58 & 61.80 \\
    & DartQuant & 68.75 & 45.04 & 71.2 & 54.8 & 73.36 & 47.78 & 75.26 & 43.60 & 77.86 & 61.96 \\
    & Ours f.p. calib. & 70.56 & 45.04 & 70.17 & 55.53 & 72.98 & 47.10 & 75.89 & 42.40 & 76.61 & 61.80 \\
    & Ours quant. calib. & 69.46 & 44.88 & 70.13 & 55.75 & 75.76 & 48.21 & 77.78 & 44.80 & 77.80 & 62.51 \\
    \midrule
    \multirow{6}{*}{4-4-4} & QuaRot & 70.4 & 44.68 & 69.11 & 53.14 & 73.02 & 44.45 & 74.63 & 41.20 & 76.71 & 60.82 \\
    & DFRot & 70.64 & 43.76 & 71.14 & 53.42 & 72.9 & 43.52 & 74.79 & 41.20 & 76.66 & 60.89 \\
    & SpinQuant & 68.03 & 45.24 & 69.45 & 52.75 & 72.35 & 47.35 & 74.34 & 41.80 & 77.26 & 60.95 \\
    & DartQuant & 68.03 & 45.39 & 70.58 & 53.35 & 73.53 & 46.76 & 74.89 & 41.0 & 77.31 & 61.20\\
    & Ours f.p. calib. & 70.96 & 44.32 & 69.75 & 54.17 & 72.81 & 46.50 & 75.16 & 41.60 & 77.31 & 61.40\\
    & Ours quant. calib. & 68.90 & 45.91 & 71.20 & 53.75 & 72.47 & 47.27 & 75.43 & 43.40 & 76.71 & 61.67\\
    \bottomrule
  \end{tabular}
\end{table}

\begin{table}
  \caption{Full perplexity results for Llama-3 70B}
  \label{tab:ppl_3_70b}
  \centering
  \begin{tabular}{clcccc}
    \toprule
    Bits (W/A/KV) & Method & WikiText2$\downarrow$      & PTB$\downarrow$ & C4$\downarrow$ & Avg$\downarrow$ \\
    \midrule
    \multirow{6}{*}{4-4-16} & QuaRot & 8.10 & 16.21 & 15.47 & 13.26  \\
    & DFRot & 7.25 & 14.26 & 13.60 & 11.70  \\
    & SpinQuant & 5.45 & 11.50 & 10.98 & 9.31 \\
    & DartQuant & 5.04 & 10.18 & 9.72 & 8.31 \\
    & Ours f.p. calib. & 4.95 & 10.19 & 9.60 & 8.25 \\
    & Ours quant. calib. & 5.09 & 10.13 & 9.65 & 8.29 \\
    \midrule
    \multirow{6}{*}{4-4-4} & QuaRot & 8.25 & 16.61 & 15.74 & 13.53  \\
    & DFRot & 7.40 & 14.70 & 13.95 & 12.02 \\
    & SpinQuant & 5.49 & 11.11 & 10.59 & 9.06 \\
    & DartQuant & 5.13 & 10.35 & 9.91 & 8.46  \\
    & Ours f.p. calib. & 5.04 & 10.39 & 9.76 & 8.40\\
    & Ours quant. calib. & 5.18 & 10.35 & 9.93 & 8.49 \\
    \bottomrule
  \end{tabular}
\end{table}

\begin{table}
\centering
\scriptsize
\setlength{\tabcolsep}{5pt} 
\renewcommand{\arraystretch}{1.08}
  \caption{Full common sense reasoning results for Llama-3 70B}
  \label{tab:common_sense_3_70b}
  \centering
  \begin{tabular}{cl ccccccccc c}
    \toprule
    Bits (W/A/KV) & Method & WG      & SIQA & LAMB & MMLU & ARC-E & ARC-C & HS & OBQA & PIQA & Avg\\
    \midrule
    \multirow{6}{*}{4-4-16} & QuaRot & 65.04 & 43.60 & 62.82 & 45.80 & 69.02 & 43.26 & 70.57 & 40.60 & 77.09 & 57.53 \\
    & DFRot & 68.75 & 43.3 & 65.11 & 50.77 & 73.7 & 49.91 & 75.28 & 42.6 & 77.09 & 60.72 \\
    & SpinQuant & 76.72 & 46.26 & 75.18 & 63.23 & 77.82 & 54.10 & 80.85 & 45.80 & 80.74 & 66.74 \\
    & DartQuant & 78.37 & 46.37 & 76.29 & 67.13 & 79.25 & 55.38 & 83.05 & 46.00 & 81.56 & 68.16 \\
    & Ours f.p. calib. & 76.48 & 47.13 & 76.98 & 67.66 & 79.97 & 54.44 & 81.70 & 47.60 & 81.56 & 68.17 \\
    & Ours quant. calib. & 76.48 & 47.85 & 75.70 & 66.09 & 78.28 & 54.86 & 82.39 & 45.00 & 81.99 & 67.63 \\
    \midrule
    \multirow{6}{*}{4-4-4} & QuaRot & 69.93 & 44.22 & 61.73 & 45.31 & 68.60 & 42.83 & 70.06 & 39.40 & 76.22 & 57.59 \\
    & DFRot & 68.82 & 42.89 & 64.53 & 47.30 & 71.72 & 48.63 & 74.37 & 41.20 & 76.88 & 59.59 \\
    & SpinQuant & 73.16 & 45.45 & 73.37 & 62.97 & 77.27 & 52.82 & 80.36 & 43.80 & 80.52 & 65.52 \\
    & DartQuant & 76.87 & 46.16 & 75.65 & 66.42 & 79.29 & 54.52 & 83.03 & 44.20 & 82.10 & 67.58\\
    & Ours f.p. calib. & 76.56 & 46.26 & 76.17 & 66.39 & 79.38 & 56.40 & 81.60 & 45.20 & 82.15 & 67.79\\
    & Ours quant. calib. & 77.43 & 48.21 & 75.41 & 66.99 & 79.04 & 56.91 & 82.17 & 45.40 & 81.61 & 68.13\\
    \bottomrule
  \end{tabular}
\end{table}

\clearpage
\newpage


\end{document}